# Scene Recognition Through Visual and Acoustic Cues Using K-Means

Sidharth Rajaram[1], J. Kenneth Salisbury[2]

[1]Monta Vista High School, [2]Stanford University

[1]sidharth.rajaram@gmail.com, [2]jks@robotics.stanford.edu

*Abstract*— We propose a K-Means based prediction system, nicknamed SERVANT (Scene Recognition Through Visual and Acoustic Cues), that is capable of recognizing environmental scenes through analysis of ambient sound and color cues. The concept and implementation originated within the Learning branch of the Intelligent Wearable Robotics Project (also known as the Third Arm project) at the Stanford Artificial Intelligence Lab-Toyota Center (SAIL-TC). The Third Arm Project focuses on the development and conceptualization of a robotic arm that can aid users in a whole array of situations: i.e. carrying a cup of coffee, holding a flashlight. Servant uses a K-Means fit-and-predict architecture to classify environmental scenes, such as that of a coffee shop or a basketball gym, using visual and auditory cues. Following such classification, Servant can recommend actions to take based on prior training.

## I. INTRODUCTION

The premise of the Third Arm Project was to develop a robotic arm along with software capabilities that could benefit developers and general consumers. For the first phase of the Third Arm, the team primarily focused on an arm that could serve as a platform for developers. This meant creating an arm that included APIs that could be leveraged to build advanced applications, ranging from mechanical tasks to learning. I specifically worked on adding capabilities to make the arm more intelligent.

The initial work for this purpose was voice control. Using the Robot Operating System (ROS) and Carnegie Mellon's CMU-Sphinx Linux speech recognition system, we constructed a voice control architecture capable of relaying instructions to servo motors through a ros-serial port. This enabled further voice control development but still did not give the arm any active intelligence. However, this served as the catalyst for Servant.

An inspiration for this idea was the thought of a situation where the third arm could offer to hold a user's coffee cup upon detecting the user was about to exit a coffee shop.

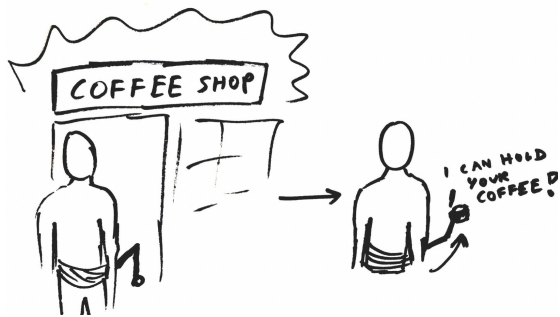

Figure 1. An artistic depiction of the vision we had in mind

This paper explores a simple architecture that uses the K-Means algorithm to identify environments through two indicators: ambient noise and perceivable color. We will focus on the implementations of Servant and example scenarios. Additionally, we will discuss limitations, lingering issues, and future opportunities.

## II. RELATED WORK

Multiple approaches to the task of acoustic scene prediction exist. Early approaches used simple algorithms such as K-nearest neighbors [4]. Most acoustic scene prediction methods were generally constructed with the end goal of providing enhanced context for the user, much in line with Servant.

Recently, however, most approaches utilize convolutional neural nets (CNNs) or at least some kind of deep learning implementation. This is largely due to the IEEE AASP Challenge on Detection and Classification of Acoustic Scenes and Events (DCASE). Most winning submissions in the DCASE Competition were built upon CNNs. It is no surprise that the usage of CNNs have resulted in acoustic scene recognition advancing as a field [1]. Most visual and acoustic classification algorithms use deep learning by default [6]. Servant aims to discover whether there is a way to accomplish the same goals by means of a simpler algorithm, namely K-Means.

## III. METHODOLOGY

In this section, we will examine how Servant works. Servant was designed entirely with Python and popular libraries such as PyAudio[1], NumPy[2], SciKit-Learn[3], OpenCV[4], and Pandas[5]. From a high-level perspective, Servant *listens and looks*. Based on what it hears and sees, it determines the environmental scene to set context for future actions. We will discuss the acoustic scene prediction method and then the visual scene prediction method.

To classify a certain acoustic scene, Servant records a five second audio file (WAV format), performs a Fast Fourier Transform (FFT) on it, and then makes predictions based on the resultant vectors. Servant trains on the resultant vectors that are products of the FFT process conducted on numerous training audio files, which are WAV files of ambient noise. In the case of Servant's current audio classification ability, the training audio files feature two distinct sounds: that of a coffee shop and that of a basketball gym.

Performing a discrete Fourier analysis on a audio file breaks the file down into its frequency components in relation to amplitude rather than time-series [5]. This enables Servant to make distinctions between audio from different types of environments; ambient noise in a coffee shop is generally different than that of a basketball gym. Refer to Fig. 2 for plots of frequencies and their respective amplitudes from a recording of a coffee shop and a basketball gym.

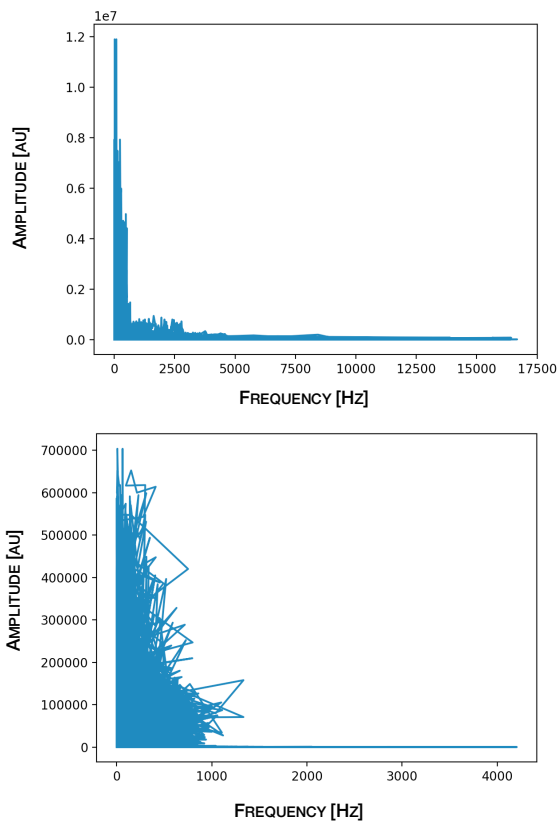

Figure 2. Above are plots of the discrete Fourier analyses of five second recordings of (1) a basketball gymnasium and (2) a coffee shop

Once an FFT analysis is performed on an audio file, a two dimensional tensor is generated. The first dimension contains the frequency components of the file while the second dimension contains the corresponding amplitude data for each frequency. Since the K-Means algorithm will classify acoustic scenes based on these tensors, the tensors will require a preprocessing step.

This step is simply creating a one dimensional array by merging the first dimension and second dimension's arrays. To do this, we append the amplitude data, [$a_1 \ldots a_n$], to the frequency data, [$f_1 \ldots f_n$]; This creates one long array. This is because the K-Means algorithm method from Sci-Kit Learn [3] can only train and predict off on one-dimensional data-points, even if they represent multidimensional information.

This preprocessing method was used anytime audio files needed to be used for training or predictions.

Acoustic scene training was done using the following steps: (1) Create a training set of **N** audio files each containing audio of the different environments desired. (2) conduct the aforementioned preprocessing method for each audio file in the training set of size **N**. This means there are now **N** one-dimensional arrays that contain frequency and amplitude data about the **N** audio files. (3) Provide **N** arrays as parameters for the fit($A_1$, $A_2$, ... $A_N$) method from the sklearn.cluster.KMeans class from the Sci-Kit Learn Python library. From this, a KMeans clustering object is created.

On a high-level, the K-Means algorithm trains by discerning clusters in the training data [2]. The number of clusters is equal to the number of unique acoustic scenes represented in the training set. Once clusters are determined, labels are assigned to each cluster. If there are **K** clusters, then the possible labels for each cluster are 0 ... (K-1). This covers the training stage.

During the prediction stage, a new audio file is presented. The preprocess steps in Fig. 3 are taken. The resultant array '**A**' of the file's frequencies and amplitudes is fed as the primary parameter to the predict(**A**...) method from sklearn.cluster.KMeans. The method determines what cluster the new file belongs to and returns the label corresponding to the target cluster.

*An example:* In the case of a training set with two distinct types of audio samples (basketball gyms and coffee shops), the number of clusters would be two; **K** = 2. The labels for the two clusters of data would be 0 and 1. Now to test the prediction ability, a test audio file of a coffee shop is presented. The file is preprocessed and the resulting audio data, '$A_C$' is provided to the predict($A_C$ ...) method. Assuming the basketball gym cluster is labeled 0 and the coffee shop cluster is labeled 1, the method would return "1." This was the extent of the K-Means based acoustic scene recognition.

Now let's take a look at the visual scene predictions of Servant. From the start, we decided that the vision aspect of Servant could be far less complex. Reason being, APIs already exist for accurate computer vision. Due to this, Servant is built to easily integrate more sophisticated vision APIs. For this initial project, the extent of Servant's computer vision is OpenCV based color sampling.

Servant takes a picture of the scene in front of it through a webcam. OpenCV is used to access the pixel values of the image. It then uses K-Means clustering to determine the most dominant colors present in the picture, as seen in Fig. 4. From there, the colors are tested against a training set of colors from pictures of different environments. This yields a prediction. Multiple pictures are taken within a 20 second window for confirmation.

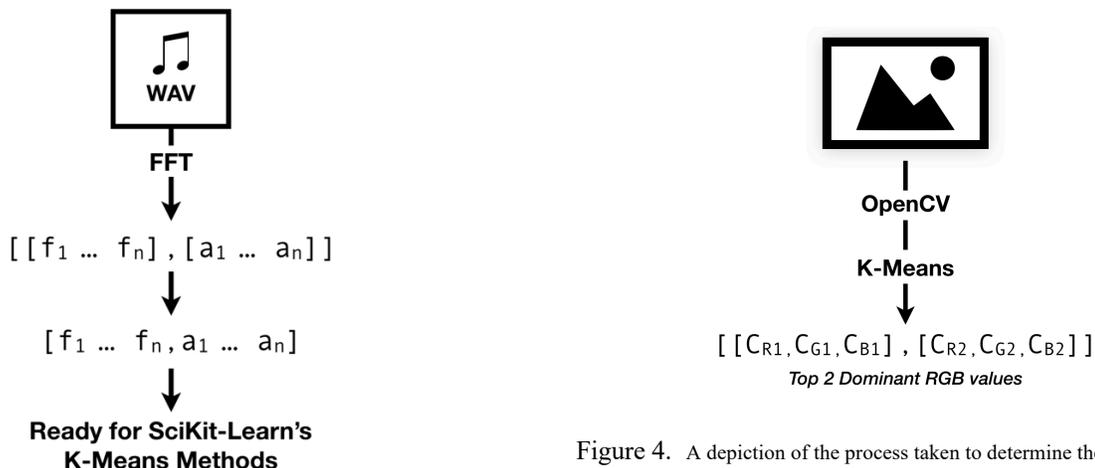

Figure 3. Above is a depiction of the process taken to prepare the FFT output for K-Means operation.

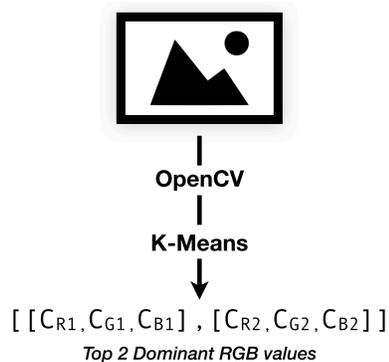

Figure 4. A depiction of the process taken to determine the most dominant colors in a given picture.

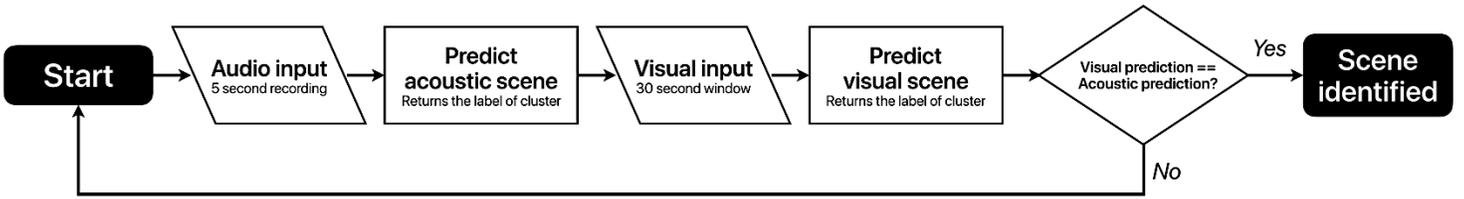

Figure 5. Above is an overarching flowchart that describes the Servant process followed to identify an environmental scene.

The final scene recognition process combined the aforementioned audio and visual prediction capabilities. It accomplishes this by comparing the two scene predictions: if the predictions are the same, then a scene has been identified. As seen in Fig. 5, the acoustic prediction occurs first. To ensure timely predictions, the visual prediction must occur within 30 seconds of the acoustic scene prediction. If no matching visual predictions have been made within that window, the process restarts. During the visual prediction phase, multiple photos are taken, to increase the confidence of a said visual prediction. Additionally, both the acoustic and visual predictions have a degree of confidence associated with each of them. This confidence metric is computed using a novel approach detailed in Section IV (Confidence).

## IV. CONFIDENCE

When it came to the acoustic and visual scene predictions, we developed a unique confidence metric specifically for K-Means. The metric is simply a scaled Euclidean distance between the test data point and the closest mean (the mean point of the cluster the test point is nearest to). Closer to 100 means more confidence.

$$C_{acoustic} = 100 - \frac{\sqrt{(A_0 - B_0)^2 + ... + (A_n - B_n)^2}}{10000} \quad (1)$$

$$C_{visual} = 100 - \sqrt{(A_0 - B_0)^2 + ... + (A_n - B_n)^2} \quad (2)$$

'A' signifies the data point being tested. 'B' signifies the closest mean. The subscripts indicate the length of each data point. Eqn. (1) includes 1/10000 to act as a scaling coefficient because, when unscaled, the acoustic error tends to be very large number even when the prediction is accurate. Even if the difference between each value in the data point is small, the sum of all the differences is very large. This is because each acoustic data point is massive in length and an Euclidean difference results in an equally massive value.

## V. ACTION LEARNING

Servant also includes methods that facilitate the prediction of actions to take once a scene has been recognized.

An example of this would be the coffee shop scenario. Once Servant has recognized it is in a coffee shop, we wanted it to trigger the servo motor to move up and consequently position the arm to hold a cup of coffee. This was accomplished by a process dubbed 'action learning.' It consists of a training period that occurs in real time. This means that the user can spend time to train the machine by themselves on what to do and then expect an accurate prediction during the testing phase.

An analogy would be a violin teacher guiding a pupil's arm with the goal of teaching the pupil how to play a note on a violin. Once the teacher has guided the pupil's arm enough, the pupil should soon figure out how to play the note by themselves.

In the real-time training phase, the user would show the desired scene to the machine and once the machine registers the scene (audio sample and visual color sample), the user would input what intended action he/she would like the machine to do. In the case of our third arm, we have been developing a "recording" mode, where the user can guide the arm to a position within a certain time window, thereby providing the various joint thetas as the intended action. So far we have not been able to produce this process in a physical domain, but we have been able to simulate the process in a script. The learning takes place by means of a simple neural net. The neural net trains on numerically encoded versions of the commands given by the user as inputs and the desired angles of motion as outputs. Fig. 6 shows terminal output during a training and testing scenario.

```
TRAINING PHASE:
Scene label: coffee
What action should I take? 42

Scene label: coffee
What action should I take? 42

Scene label: coffee
What action should I take? 42

Scene label: gym
What action should I take? 10

Scene label: gym
What action should I take? 10

Scene label: coffee
What action should I take? 42

Scene label: gym
What action should I take? 10

output layer error after 0 iterations: 0.5977810570699277
output layer error after 1000 iterations: 0.012277542481386563
output layer error after 2000 iterations: 0.006025536372095795
output layer error after 3000 iterations: 0.004272927593187904
...
output layer error after 97000 iterations: 0.00047381829512478435
output layer error after 98000 iterations: 0.00047110709909633657
output layer error after 99000 iterations: 0.0004684395456174968

PREDICTION PHASE:
Tell me something: coffee
based on your command, here's my action prediction:
[['42']]

Tell me something: gym
based on your command, here's my action prediction:
[['10']]
```

Figure 6. Simulation of the real-time training system. Training inputs are the labels of scenes. Training outputs are the intended angles of motion for the arm.

## VI. RESULTS

To test Servant, we utilized a standard microphone and a webcam to act as the input devices for the acoustic and visual prediction methods, respectively. We would play an audio file of an environment (coffee shop ambient noise or basketball gym sounds) and show the webcam an image of an environment. Sometimes we would show and play matching images to test whether Servant could identify a coherent scene. Other times we would provide mismatching audio files and images in order to test Servant's ability to identify inputs as that of different scenes. Below are the summarized results of various test scenarios we ran Servant through. However, more detailed results and exact terminal outputs are located at: http://sidrajaram.me/servantoutput

TABLE I.   TERMINAL OUTPUT BASED ON TESTED STIMULI

| Visual stimuli | Audio stimuli | |
|---|---|---|
| | *Coffee shop recording* | *Basketball gym recording* |
| Coffee shop picture | `CoffeeScene detected` | `No scene detected` |
| Basketball gym picture | `No scene detected` | `GymScene detected` |

TABLE II.   AVERAGE CONFIDENCE METRICS FOR TESTED STIMULI

| Visual stimuli | Audio stimuli | |
|---|---|---|
| | *Coffee shop recording* | *Basketball gym recording* |
| Coffee shop | 87.94 | 0.000 |
| Basketball gym | 0.000 | 65.38 |

Table 1 shows the result matrix of Servant's test performance (Terminal output shows detected scene, if any). Additionally, we determined the confidence metric for each scenario. Table 2 includes the corresponding average confidence values (average of visual and acoustic confidences) for the scenarios in Table 1.

## VII. CONCLUSIONS

Servant's results proved to us that it is possible to accomplish rudimentary scene recognition with an algorithm as simple as K-Means. Due to the results of deep-learning-based solutions at DCASE, we look forward to potentially incorporating neural nets in conjunction with our K-Means implementation.

The results also illuminated obstacles inherent to acoustic sampling. Making sure the signals from the acoustic samples were ergodic was a challenge. For example, in some cases, a five second recording at a coffee shop may include a noise generally considered foreign to a coffee shop. This occasionally skewed results. Our response to this issue was simply adding more audio clips to the training dataset. Another idea was to make Servant capable of asking for help if absolutely necessary. When Servant becomes robust enough to be utilized in mission critical applications like disaster relief, it may sometimes be beneficial for the user if Servant asks them if its prediction is correct.

Another big takeaway from the work with Servant was that there exists a relatively simple method of audio classification. Using off-the-shelf Fourier analysis methods and a unique preprocessing method as mentioned in Section III (Methodology), we were able to discern different audio types.

A mainstay of the future development of Servant will be making the acoustic and visual analysis happen in parallel rather than sequentially. Additionally, there may be other characteristics to analyze to make a scene recognition. Since characteristics like dominant colors might not be the most reliable in real world scenarios, we hope to begin looking at other distinct environmental features such as motion, temperature, altitude, etc. Fig. 7 displays how a future decision path may look like.

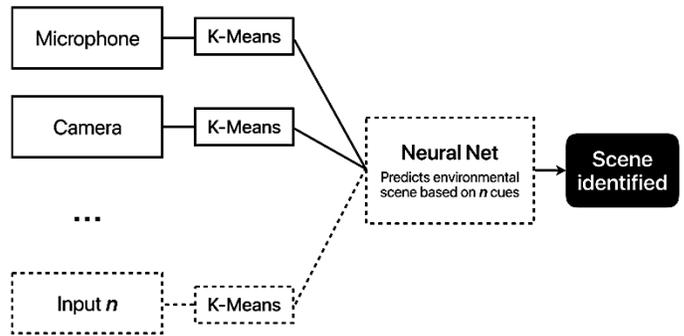

Figure 7. Objects with dashed borders indicate future developments. We hope to analyze a variety of inputs such as temperature and acceleration. Our K-Means methods will indicate the scene predictions based on each individual input and a newly added neural net will be able to develop a cohesive scene prediction based on the K-Means methods' outputs.

We expect Servant and extensions of this work to continue making the Third Arm more intelligent. A machine capable of conducting scene analysis and carrying out contextual actions will have numerous real-world applications and cascading benefits for user experience.

## VIII. GLOSSARY

This serves as a section to define some terms mentioned earlier in the paper.

1. PyAudio: Python bindings for audio I/O. Used to play and record audio on a variety of devices/platforms.

2. NumPy: Python library to support multidimensional array operations and high-level mathematical functions.

3. SciKit-Learn: Python library for machine learning.

4. OpenCV: Library built for computer vision

5. Pandas: Python library for data analysis and manipulation.